# Role of Intonation in Scoring Spoken English

Amber Nigam
kydots.ai
Delhi, India
amber@kydots.ai

Arpan Saxena
kydots.ai
Delhi, India
arpan@kydots.ai

Ishan Sodhi
IIT Bombay
Delhi, India

*Abstract* – In this paper, we have introduced and evaluated intonation-based feature for scoring the English speech of non-native English speakers in Indian context. For this, we created an automated spoken English scoring engine to learn from the manual evaluation of spoken English. This involved using an existing Automatic Speech Recognition (ASR) engine to convert the speech to text. Thereafter, macro features like accuracy, fluency and prosodic features were used to build a scoring model. In the process, we introduced SimIntonation, short for similarity between spoken intonation pattern and "ideal" i.e. training intonation pattern. Our results show that it is a highly predictive feature under controlled environment. We also categorized inter-word pauses into 4 distinct types for a granular evaluation of pauses and their impact on speech evaluation. Moreover, we took steps to moderate test difficulty through its evaluation across parameters like difficult word count, average sentence readability and lexical density. Our results show that macro features like accuracy and intonation, and micro features like pause-topography are strongly predictive. The scoring of spoken English is not within the purview of this paper.

## I. Introduction

Scoring spoken English has gained popularity in the recent years and is being used in various academic and industrial assessments. The need of the scoring is derived from the fact that English is a widely spoken and recognized language. It is often a prerequisite for many jobs and academic programs. However, scoring proficiency of spoken English is not a simple process and requires simultaneous assessment of a lot of features of spoken English, simple and complex alike.

The first step to scoring spoken English is having a good Automatic Speech Recognition (ASR) system. It is the technology that allows a computer to recognize and convert into text, spoken words. Developing an ASR system that can recognize, and thus convert all speech (spoken by anyone) to text is a huge challenge that requires tremendous amounts of data.

Then output of ASR along with a few speech features are used to score using a model. Some of these features include, but are not limited to fluency, accuracy, and intonation. Machine learning and deep learning algorithms can then be used to score the spoken English after learning from manual evaluation of the training samples.

It is also important that automated evaluation accounts for the influence of background of the speaker being evaluated. For example, a large segment of people in some parts of India replace phoneme '/t' with phoneme '/th' while pronouncing some proper nouns like the name 'Kartik', and this is uniformly acceptable in the region. The automated assessment can address such nuances through a well-balanced selection of the training dataset.

## II. Related Work

Human test scorers have a wide variety of problems associated with them. First is the problem of the time taken and the cost involved.

Second, is the very fact that they're human and thus prone to a degree of bias associated with them with respect to vocabulary, understanding, to name a few. Even someone's consistency in scoring assignments is worth mentioning, especially when there are a large number of assignments. Then there's the issue of different people having different standards of considering something as average.

Two human scorers are likely to grade the same language assignment differently, based on whether it's below or above their considered view of average. All these issues are well described in (Zhen Wang and Alina A von Davier, 2014).

The existence of problems described above led to exploring ways of automating the entire scoring process. An automatic scoring system is only as good as the features of human scoring used as the basis for constructing the model. Broadly speaking, the automated scoring systems involve an additional 2 phase evaluations, after the initial ASR phase. First, they engineer certain human devised characteristics that can be mapped to certain aspects of language proficiency. Then in the next phase, using supervised learning algorithms, the system can process those characteristics, on the basis of human-scores to train it (Zhou Yu et al., 2015). After the training phase, the system is ready for testing.

Furthermore, Deep Neural Networks can be used to combine low-level characteristics to generate high-level characteristics, without any human intervention. This would richly contribute to earlier scoring models, which are heavily dependent on human-engineered characteristics. For instance, the Bidirectional Long Short-Term Memory (Bi-LSTM) could be used to join different characteristics for scoring spoken a constructed response (Zhou Yu et al., 2015). Such an approach to induce features using Deep Neural Networks has also been successful in object recognition (Alex Krizhevsky et al., 2012) and multimodal analysis (Jiquan Ngiam et al., 2011).

Work has also been done to automatically assess language proficiency in spontaneous speech, i.e. when the spoken text is not known in advance. This can be engineered using test takers' answers to complex test items. (Klaus Zechner and Isaac I. Bejar, 2006).

Another work (Evanini, K. et al., 2017) states that speech can be evaluated very effectively, if the response content is predictable, as in the case of reading aloud a piece of text (Evanini et al. 2015) or repeating a pre-recorded speech (Cheng et al., 2014). The evaluated scores, in this case, confer with the inter-human agreement standards. Even for the questions where a sentence needs to be completed according to a template (Zechner et al., 2014). On the other hand, according to this paper (Evanini, K. et al., 2017), work in the field of evaluating questions that judge the opinion of a test-taker (Xie, 2012), is still developing i.e. it does not confer with the inter-human agreement standards. Similarly, for the questions where the response has to be a story narration, with the help of pictures, or simply retelling the story in one's own words (Somasundaran, et al., 2015). Even the questions that ask for a summary of key points (Xiong et al., 2013) with/without the help of pictures come under the above type of developing research. Taking this a step further, if the response is made more unpredictable, like asking the test-taker to participate in an open dialogue with an imaginary interlocutor, the chances of a inter-human agreement on scores reduce drastically.

Moreover, the papers like (Evanini, K. et al., 2017) and (Breyer et al., 2017) describe hybrid approaches of using human and automated scores that can deliver a better performance than using either individually. Finally, work is also being done in solving problems by collaborating (Bassoon et al., 2016) and interactive speech (Evanini et al., 2014). This would extend the influence of Natural Language Processing in formative assessments and learning applications, in time to come.

## III. AUTOMATIC SPEECH RECOGNITION

An automatic speech recognition (ASR) module forms the basis of almost all the spoken Language evaluation systems, mentioned above. According to this paper (Zhou Yu et al., 2015) an ASR front-end module for most state-of-the-art evaluation systems gives word conjectures about the replies given by the person appearing for the test. Therefore, it can be clearly anticipated that a large amount of data, more specifically a collection of non-native speech, and exact transcriptions of each piece of that speech, would be needed to train this type of ASR module. Needless to say, this would involve human effort in transcribing the entire speech collection.

The ASR system used in this work is Sphinx4, a popular, open source Java speech recognition system. It has a comprehensive list of features that can be tuned to improve ASR's performance (see Table 1 for the current configuration).

TABLE I. ASR FEATURES

| ASR Feature | Value |
|---|---|
| Lowerf | 130 |
| Upperf | 6800 |
| Nfilt | 40 |
| Transform | Dct |
| Lifter | 22 |
| Feat | 1s_c_d_dd |
| Agc | None |
| Cmn | Current |
| Varnorm | No |
| Cmninit | 40,3,-1 |

## IV. EXPERIMENT

This section describes the experiments conducted in order to score spoken English of nonnative speakers. A set of 100 people was chosen from relatively diverse backgrounds for gathering data that would be used. Each person read aloud the same set of sentences in a curated environment. No prior knowledge regarding the experiment was circulated beforehand, so as to avoid any kind of preparations for the same.

The sets of sentences given to the users were validated against the following metrics (Nigam, A, 2017) to ensure uniformity in difficulty levels of the sets: (a) Number of words having more than 4 syllables (b) Lexical Density (Ure, J, 1971) (c) Readability (Kincaid JP, 1975). Furthermore, "bad" sentences (Yuan, J. et al., 2015) were removed from the set of sentences presented to the candidates.

We used crowdsourcing to train our model through an organization. Each sound file fed to the ASR engine, Sphinx4, is a 16 kHz 16 bit mono WAV file. We tuned ASR features like lower frequency, upper frequency, number of filters used and direct cosine transform. Some of the important configuration values for the ASR system are given in the ASR section. The output of the ASR is transcription of the spoken text along with metadata like confidence value, which is used for scoring.

Each candidate was assigned a score between 1 and 6 both manually and through automatic system. The dataset was divided into approximately 60:20:20 among training, validation, and testing sets while using machine learning and deep learning algorithms. The manual scoring was done by two English speakers who are proficient in English language. They had an inter-rater agreement of .83. The mean score and the standard deviation were 4.3 and 1.3 respectively. We used SimIntonation to quantify the intonation based characteristics of the spoken English. Our results show that SimIntonation is a predictive feature and validated it with Student's t-test at a significance level of 0.05. We also took steps to evaluate SimIntonation's correlation with other features and did not find any it significantly correlated with any other feature.

## V. FEATURES

The feature set comprises macro features like accuracy, fluency, and intonation. Most of these features are used in practice today by human scorers.

Accuracy: It is a simple feature that evaluates the similarity between the ASR's transcription of the spoken text and the text that was supposed to be spoken by the candidate. We evaluate the similarity using Levenshtein distance algorithm at word level.

Fluency: Fluency can be described as the flow of speech. We quantify fluency through following micro features like average speed or words per minute, articulation rate, and pauses. Articulation rate is a rate of speaking in which all pauses are excluded from the calculation. Our findings are in sync with this paper (Dankovičová, 1999) that shows that the variation in articulation rate is of importance for speech evaluation. Pauses have been categorized into two kinds based on the con-tent: (a) filled pauses and (b) unfilled pauses. Based on their length, pauses have been categorized into: (a) short pauses and (b) long pauses.

Intonation: It is the rise and fall of the voice in speaking. For example, a declarative sentence "Adam wants to walk but Eve prefers to take a taxi." can be broadly divided into 2 parts with respect to intonation, termed as Intonational Phrases (Nespor and Vogel, 1987). The first phrase begins with a high tone and then reduces in tone, after which it ends with an increasing tone. On the other hand, the second one finishes with a falling tone—implying the end of a complete utterance. An interrogative statement starts low and ends high, unlike the Declarative sentence. As the intonation becomes flat or grows closer to a monotone, its quality reduces.

We evaluated intonation through SimIntonation that is the correlation between pitch values for words of a testing sentence and pitch values for words of the corresponding training sentence (average pitch values for words is taken in case a sentence is trained by multiple people). It quantifies an intonation similarity between supposedly ideal training intonation pattern and the spoken intonation pattern. We found this feature to be significantly predictive of manual scoring under con-trolled training environment.

## VI. RESULTS

In this paper, we have shown that SimIntonation is a predictive feature of manual scoring at a significance level of 0.05. Besides, it is not a derivative or correlated feature of other features that we considered. We took steps to ensure that testing is not biased by difficulty level of test furnished to candidates, testing environment conditions like background noise, or biased training dataset due to difference in background of people whose voices were used in training and testing datasets.

## VII. CONCLUSION AND FUTURE WORK

In this paper, we introduced and evaluated intonation-based feature for scoring spoken English of nonnative speakers. We introduced a way to quantify similarity between testing intonation with the training intonation through SimIntonation that was strongly predictive for our dataset. We categorized pauses on the basis of duration (long and short pauses) and their content (filled and silent pauses) to study the impact in greater depth. We also took efforts like having a hand-picked dataset for training Automatic Speech Recognition (ASR) to recognize subtleties in nonnative speech better.

Our next step is to create a spoken English scoring Engine that uses the features identified in this paper. We also plan to test it on people be-longing to different backgrounds and to map the effect of background with the subtleties in spoken English and its effect on manual scoring.

## ACKNOWLEDGMENT

We would like to acknowledge that this research is supported by kydots.ai. We would also like to thank all the people who crowdsourced to provide us the all-important data on which the experiments were conducted.